\pdfoutput=1

\documentclass[11pt]{article}

\usepackage{acl}
\usepackage{amsmath}
\usepackage{times}
\usepackage{latexsym}
\usepackage{booktabs}
\usepackage{multirow}
\usepackage[T1]{fontenc}

\usepackage[utf8]{inputenc}
\usepackage{amsmath}
\usepackage{microtype}

\usepackage{inconsolata}

\usepackage{graphicx}
\usepackage{enumitem}
\usepackage[utf8]{inputenc}
\newcommand{\affiliations}[1]{\def\@affiliations{#1}}
\newcommand{\ours}{\text{CKnowEdit}}

\title{\ours: A New Chinese Knowledge Editing Dataset for\\Linguistics, Facts, and Logic Error Correction in LLMs}

\author{
  Jizhan Fang\textsuperscript{1}\thanks{Equal contribution and shared co-first authorship.} \hspace{1em} 
  Tianhe Lu\textsuperscript{1,}\footnotemark[1] \hspace{1em} 
  Yunzhi Yao\textsuperscript{1} \\
  \textbf{Ziyan Jiang}\textsuperscript{1} \hspace{1em} 
  \textbf{Xin Xu}\textsuperscript{2} \hspace{1em}
  \textbf{Huajun Chen}\textsuperscript{1}\hspace{1em}
 \textbf{Ningyu Zhang}\textsuperscript{1}\thanks{Corresponding Author.} 
  \\
  \textsuperscript{1}Zhejiang University \hspace{2em} 
  \textsuperscript{2}University of California, San Diego \\
  \texttt{\{fangjizhan, yyztodd, zhangningyu\}@zju.edu.cn} \quad 
  \texttt{xinxucs@ucsd.edu}
}

\begin{document}
\maketitle
\begin{abstract}
Chinese, as a linguistic system rich in depth and complexity, is characterized by distinctive elements such as ancient poetry, proverbs, idioms, and other cultural constructs. However, current Large Language Models (LLMs) face limitations in these specialized domains, highlighting the need for the development of comprehensive datasets that can assess, continuously update, and progressively improve these culturally-grounded linguistic competencies through targeted training optimizations. To address this gap, we introduce \textbf{CKnowEdit}, the first-ever Chinese knowledge editing dataset designed to correct linguistic, factual, and logical errors in LLMs. We collect seven types of knowledge from a wide range of sources, including classical texts, idioms, and content from Baidu Tieba Ruozhiba, taking into account the unique polyphony, antithesis, and logical structures inherent in the Chinese language. By analyzing this dataset, we highlight the challenges current LLMs face in mastering Chinese. Furthermore, our evaluation of state-of-the-art knowledge editing techniques reveals opportunities to advance the correction of Chinese knowledge\footnote{Code and dataset are available at \url{https://github.com/zjunlp/EasyEdit}.}.
\end{abstract}

\begin{figure*}[ht]
    \centering
    \includegraphics[width=1\textwidth]{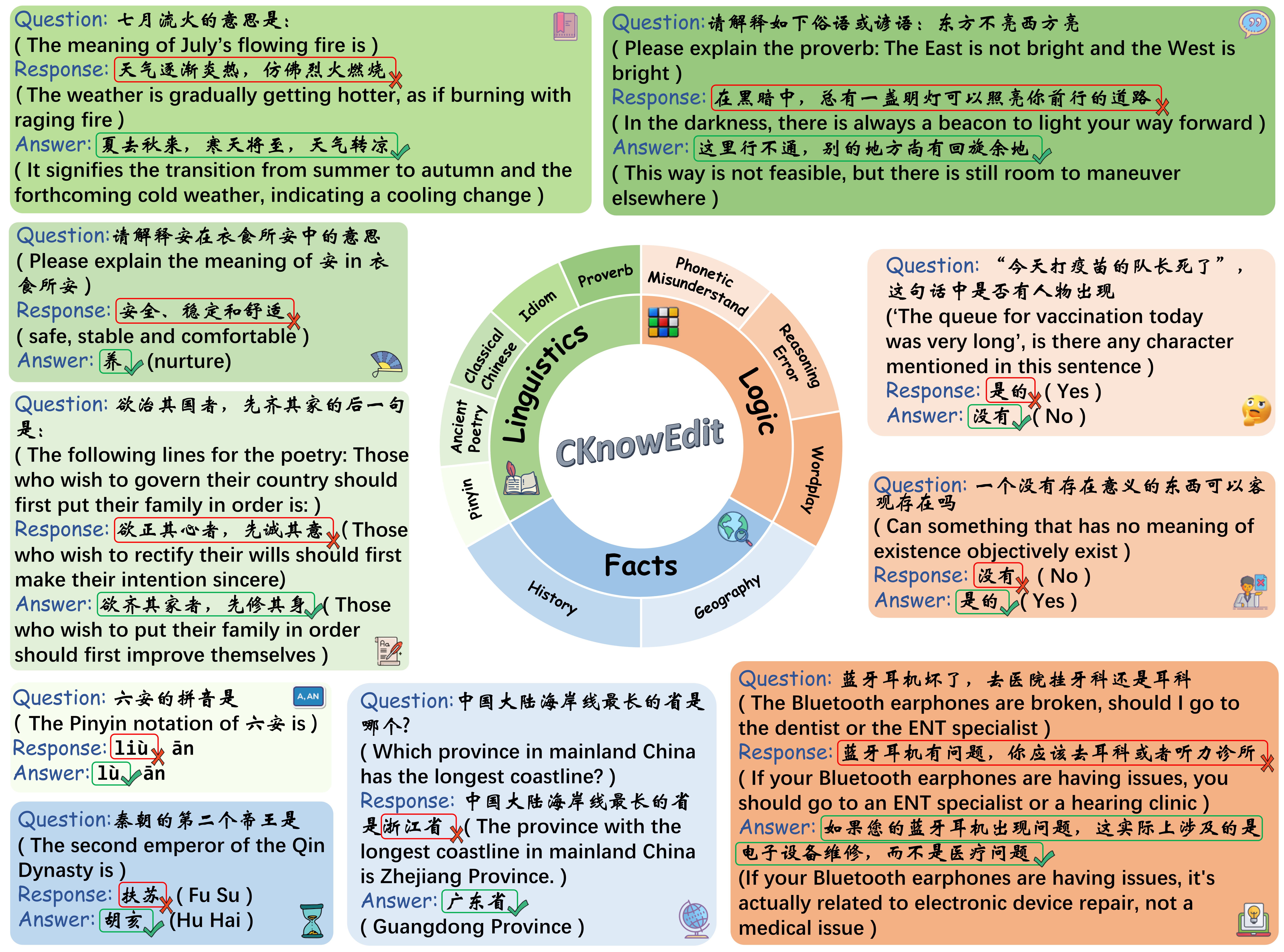}
    \caption{Examples of data from each subcategory in CKnowEdit, with detailed explanations  provided in \S\ref{sec:motivation}.}
    \label{fig:samples}
\end{figure*}

\section{Introduction}

The reliance on static training data and the lack of explicit knowledge representation in Large Language Models often lead to issues such as hallucinations, bias, and offensive outputs \cite{DBLP:journals/corr/abs-2303-18223,DBLP:journals/corr/abs-2311-05232,liu2023trustworthy,sun2024trustllm,Chen_2024}. 
These limitations become particularly pronounced when LLMs operate in complex domains or languages, such as Chinese. 
As shown in Figure \ref{fig:samples}, Chinese is a highly complex and linguistically unique system and presents three distinct challenges for LLMs versus Indo-European languages \cite{luelsdorff1994prague,matthiessen2023system,xu2023superclue,wen2023improving}: 
$(i)$ \textbf{Linguistic Complexity}: Characters intricately blend shape, sound, and meaning through composition and contextual pronunciation shifts, while flexible grammar and cultural elements (poetry, idioms, etc.) evolved over millennia. 
$(ii)$ \textbf{Culture-Laden Facts}: Untranslatable contexts in specific facts like geographical/historical terms. 
$(iii)$ \textbf{Language-Specific Logic}: Context-dependent reasoning patterns that rely on implicit connectors and topic prominence over subject-predicate structures, often leading to misalignment in logical chain extraction.

In this work, we propose to correct Chinese knowledge errors in LLMs via knowledge editing \cite{DBLP:conf/emnlp/YaoWT0LDC023,wang2023knowledge,zhang2024comprehensive,DBLP:conf/aaai/HuLHZLZ24,DBLP:journals/corr/abs-2311-08011,DBLP:journals/corr/abs-2402-13048,DBLP:journals/corr/abs-2401-10471,DBLP:conf/nips/PadmanabhanOZDC23,DBLP:conf/acl/QiaoLN24,DBLP:journals/corr/abs-2407-20224,DBLP:conf/iclr/LiLCZLW0024,hase2024fundamental,DBLP:conf/emnlp/WuPWL24,DBLP:journals/corr/abs-2401-10471}. 
Nevertheless, current research on knowledge editing predominantly concentrates on English-language factual knowledge~\cite{DBLP:conf/emnlp/CaoAT21,meng2022locating,wu2024updating}, derived from Wikipedia, which introduces an Anglo-centric bias. 
Recently, there have been some multilingual datasets~\cite{wang2023cross, xie2024memla, wei2024mlake, nie2024bmike} attempting to explore editing methods for different languages. 
However, these datasets are often created by translating the English corpus into another language, and translation \cite{vanmassenhove2019lost,berman2021translation} has been shown failing to capture the intricate linguistic features and cultural nuances inherent to special language, resulting in a loss of lexical richness and diversity. 
Meanwhile, these works are primarily designed to assess the coherence of current editing methods between different languages and are not suitable for research on language-specific (a.k.a. Chinese) knowledge editing methods or for understanding LLMs’ representation of specific languages.

To help address the three major challenges mentioned above and mitigate some existing deficiencies in the current editing datasets, we construct a new Chinese dataset, \textbf{CKnowEdit}, which takes into account language-specific characteristics, ensuring that the data is not only linguistically accurate but also culturally matched. 
To ensure the quality and diversity of CKnowEdit, we collect data from a variety of sources, including classical literature, modern colloquialisms, and Baidu Tieba Ruozhiba~\cite{bai2024coigcqia} (a popular Chinese online forum renowned for its abundance of logic puzzles and brainteasers, highly suitable for evaluating the reasoning capabilities). 
As a result, we organize CKnowEdit into 3 major categories, including \emph{Linguistic}, \emph{Facts} and \emph{Logic} corresponding to the three major challenges and 10 subcategories, as shown in Figure \ref{fig:samples}.

To benchmark the effectiveness of knowledge editing methods on CKnowEdit, we evaluate five representative methods on four models. 
Departing from traditional knowledge editing evaluations that rely on token/logit-level measurements through teacher-forcing automation~\cite{DBLP:conf/emnlp/YaoWT0LDC023}, we implement \textbf{open-ended text generation} to evaluate edited models under more realistic and demanding conditions and utillize LLM-as-a-judge paradiam to effectively evaluate. 
The results demonstrate the challenges presented by the dataset and underscore the need for more sophisticated Chinese knowledge editing approaches in the future.
Our major contributions are as follows:

\begin{itemize}
    \item We propose a new knowledge editing dataset, CKnowEdit, which is uniquely characterized by its Chinese linguistic features and cultural depth, comprehensively exploring the language's distinctiveness and the challenges it poses to LLMs from three perspectives.
    \item We report the empirical results of recent knowledge editing baselines on CKnowEdit, revealing their limitations when applied to Chinese literature.
    \item We further explore the challenges of Chinese knowledge editing and the struggles faced by existing models in understanding Chinese language and culture. 
\end{itemize}

\section{Criteria for Knowledge Sourcing}
\label{sec:motivation}

\subsection{Chinese Linguistics}

Chinese linguistics studies the phonetics, vocabulary, semantics and grammar of the Chinese language, the linguistic knowledge in CKnowEdit is categorized into five subtypes. Each subtype of knowledge presents unique challenges for LLMs.

\paragraph{Pinyin}
\emph{Pinyin Notation} serves as the official romanization system for Standard Mandarin Chinese, utilizing the Latin alphabet to represent Chinese characters phonetically. 
In Chinese, the phenomenon of polyphonic characters is widespread. 
As shown in Figure \ref{fig:samples}, the character ‘\includegraphics[width=0.35cm]{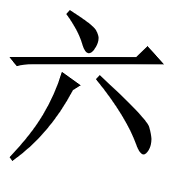}’ (six) is pronounced ‘Liù' in most cases, but in ‘\includegraphics[width=0.7cm]{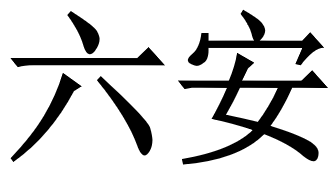}’ (a city) it is pronounced ‘Lù.' 
This inherent ambiguity in grapheme-phoneme mapping poses challenges for LLMs, especially when dealing with low-frequency characters with multiple pronunciations, which are also included in CKnowEdit.

\begin{figure*}[ht]
    \centering
    \includegraphics[width=1\textwidth]{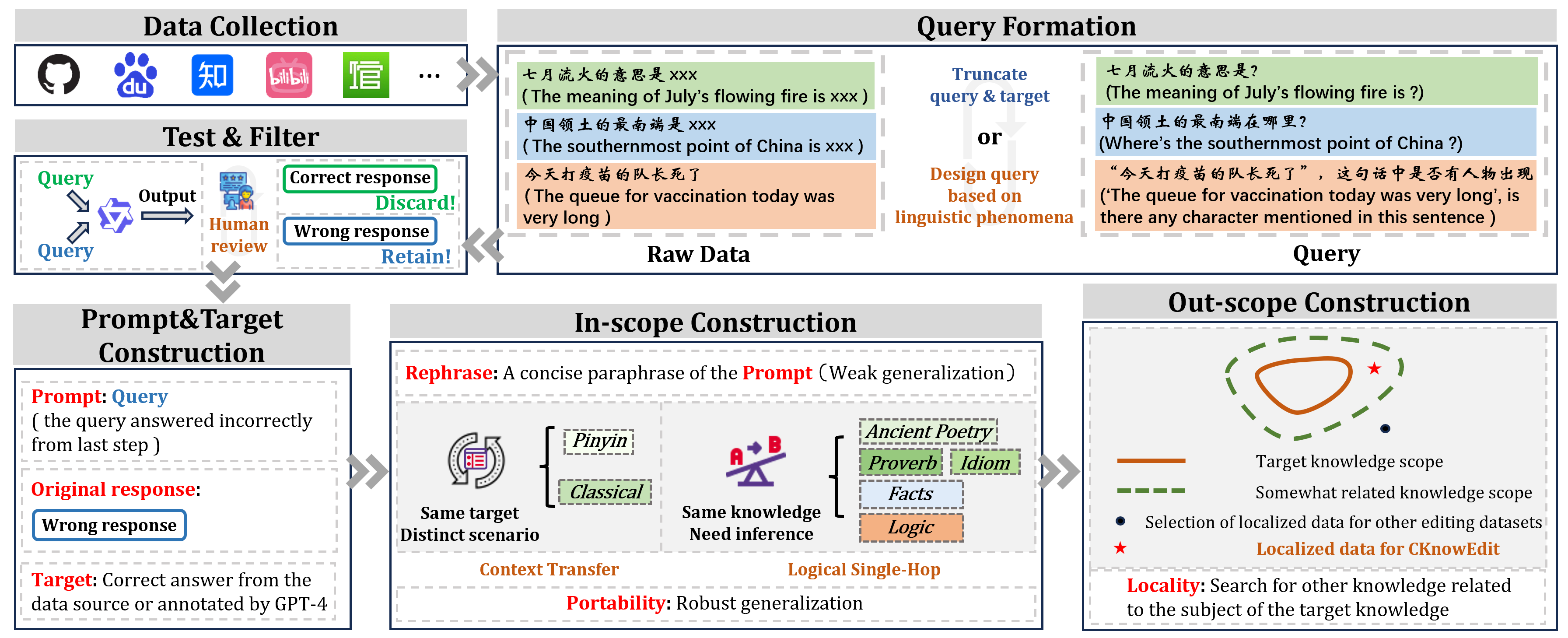}
    \caption{Overview of \ours\ construction. A full sample of CKnowEdit is shown in Figure \ref{fig:case_1} and \ref{fig:case_2}.}
    \label{fig:overview}
\end{figure*}

\begin{figure}[ht]
    \centering
    \includegraphics[width=1\linewidth]{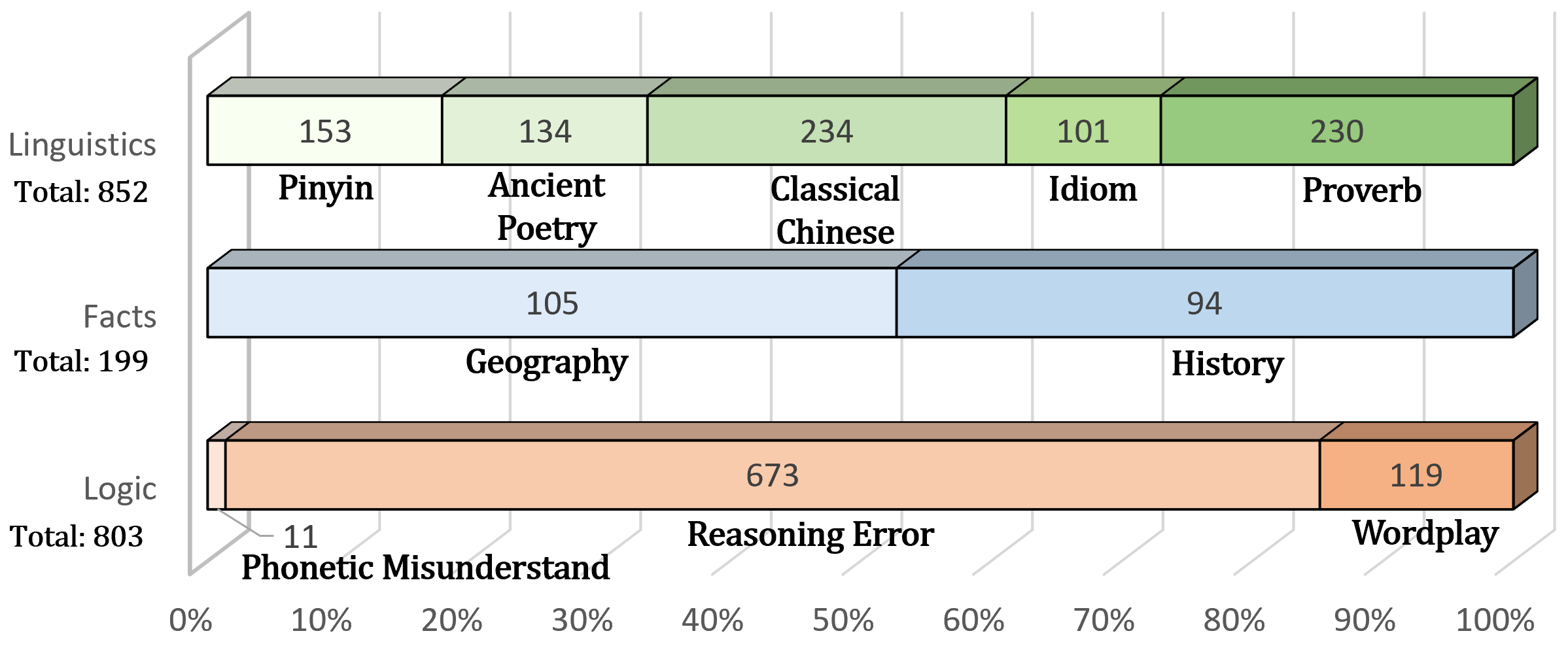}
    \caption{The statistics of CKnowEdit.}
    \label{fig:statistics}
\end{figure}

\paragraph{Ancient Poetry}
\emph{Ancient Poetry} constitutes an essential component of Chinese classical literature, which significantly differs from Modern Vernacular Chinese, particularly in semantic constructs and graphological conventions.
Additionally, \emph{Ancient Poetry} adhere to extremely strict requirements for format and rhythm, where every character must be precise and cannot be altered or omitted.
This form of ancient language commonly embedded in the parameters of large language models, poses a significant challenge to their memory and processing capabilities.

\paragraph{Classical Chinese}
Words in \emph{Classical Chinese} often carry greatly different meanings compared to Modern Chinese. 
And the same character may represent distinct concepts based on context.
As shown in Figure \ref{fig:samples}, the ‘\includegraphics[width=0.35cm]{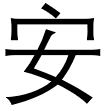}’ ( means ‘safety’ in Modern Chinese ) can denote ‘to nurture’, ‘to stabilize’ or function as an interrogative term (‘where/how’) in classical texts. 
This semantic divergence poses unique challenges for language models trained on Modern Chinese data, particularly when processing context-sensitive interpretations of polysemous characters in classical literature.

\paragraph{Idiom}
Directly comprehending Chinese \emph{idioms} or interpreting them literally often leads to a loss of their true meaning. 
In fact, the actual meaning of many idioms can be entirely opposite to their literal interpretation, such as the idiom \includegraphics[width=1.3cm]{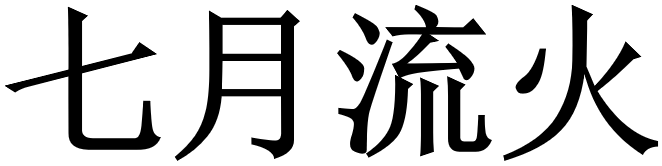} which literal meaning is July’s flowing fire contrary to the true meaning. 
LLMs' statistical learning paradigms struggle to resolve these interpretative gaps, particularly when processing idioms whose surface forms actively contradict their established semantic values in linguistic praxis.

\paragraph{Proverb}
\emph{Proverbs} often use modern expressions with clear literal meanings, but their actual significance usually depends on metaphorical understanding. While these proverbs maintain consistent core meanings, LLMs struggle to apply them appropriately across different real-life situations.

\subsection{Factual Knowledge}

\emph{History} and \emph{Geographical} knowledge in CKnowEdit covers key events and historical figures, regional landscapes, and unique local cultures across China. 
However, mainstream LLMs demonstrate notable gaps in their understanding of factual knowledge related to China’s history and geography~\citep{sun2024benchmarking}.

\subsection{Chinese language-specific logic trap}
\paragraph{Phonetic Misunderstand}
Figure \ref{fig:samples} demonstrates a typical Chinese \emph{phonetic misunderstanding} involving the polyphonic character  ‘\includegraphics[width=0.35cm]{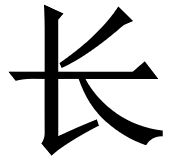}’. 
When pronounced as ‘zhǎng’, it combines with the preceding ‘\includegraphics[width=0.35cm]{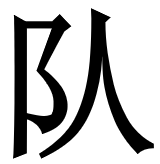}’ to form ‘\includegraphics[width=0.7cm]{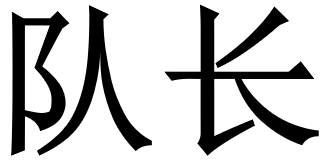}’ (team leader), suggesting the illogical meaning ‘The vaccinated team leader has died’. 
However, ‘\includegraphics[width=0.35cm]{figures/chang.png}’ actually functions as an adjective meaning ‘was long’ which pronounced as ‘cháng’, and ‘\includegraphics[width=0.35cm]{figures/dui.png}’ simply means ‘queue", indicating that ‘Today's vaccination queue was extremely long’. 
This highlights how LLMs' pronunciation disambiguation failures can lead to semantic misinterpretations, even with proper word segmentation.

\paragraph{Reasoning Error}
When meeting complex reasoning tasks in the Chinese language, LLMs may commit \emph{reasoning errors}, hence CKnowEdit has incorporated such data into its considerations.

\paragraph{Wordplay}
This type of logical fallacy often arises from word segmentation errors or ambiguous terms being misinterpreted as unintended meanings, thereby distorting the original semantic content of the textual components within a sentence. 
As illustrated in Example 1, the LLM misinterpreted \includegraphics[width=1.40cm]{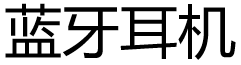} (Bluetooth earphones) through erroneous word segmentation as ‘\includegraphics[width=1.00cm]{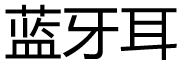}-\includegraphics[width=0.35cm]{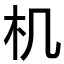}’ (literally ‘blue tooth-ear device’), forcing a literal interpretation within a physiological context (teeth and ears), ultimately producing semantically absurd outputs.

\section{The Construction of CKnowEdit}


\subsection{Data Preprocess}

\paragraph{Data Collection}
As described in \S\ref{sec:motivation}, we classify the data types in CKnowEdit into 3 major categories and 10 subcategories, as also illustrated in Figure \ref{fig:samples}. 
Data collection is conducted based on this classification.
We crawled authentic and diverse Chinese corpora and collected 11,981 raw data entries initially. 
All our data collectors adhere to the copyright and licensing terms of each data source website and all the data collected are freely available for academic research. 
Detailed addresses of each source website can be found in Appendix \ref{sec:source_web}.

\paragraph{Data Filtering}
As shown in Figure \ref{fig:overview}, we first convert all the collected raw data into queries and pose them to the Qwen-7b-chat \cite{qwen2} model as a baseline. 
We then retain only those questions that the model answered incorrectly, discarding those it answered correctly. 
The filtering process ensures CKnowEdit remains challenging and justifies the necessity of applying knowledge editing techniques. 
To maintain data quality, we conduct a manual review of all collected responses.

\subsection{Data Annotation}

\paragraph{Prompt-target Construction}
The queries after filtering are used as the prompt field in the data. 
But for target field: fixed/data-provided answers are used directly; open-ended explanations (e.g., interpret logic errors) are generated by GPT-4. 
Moreover, all answers generated by GPT-4 undergo meticulous manual review and correction to ensure their accuracy. 

\paragraph{In-scope Construction}
Effective model editing requires consistent behavioral adjustments across all examples within the editing scope. 
Beyond correcting the primary target knowledge, related in-scope information conveying similar concepts should also be updated. 
We therefore assess two distinct generalization capabilities: weak and robust generalization. 
Specifically, we evaluate the weak generalization effect by rephrasing the prompt, such as rephrasing \textit{`Please complete the following ancient poem'} as \textit{`The next line of the following ancient poem is...'}. 
Robust generalization is measured through two approaches: 
$(i)$ \textbf{Context Transfer}: 
This involves transferring the same knowledge or language pattern to a different application scenario to see if the edited model has truly learned the knowledge. 
For example, in classical Chinese, the character ‘\includegraphics[width=0.3cm]{figures/an.png}’ means ‘nurture’ in the phrase ‘\includegraphics[width=1.2cm]{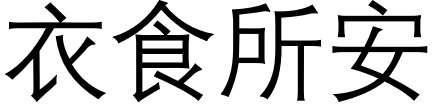}’ (provides sustenance). We then ask the edited model about the meaning of ‘\includegraphics[width=0.3cm]{figures/an.png}’ in ‘\includegraphics[width=1.2cm]{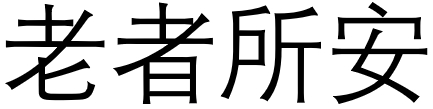}’ (the elderly are supported) where it still means ‘nurture’.
$(ii)$ \textbf{Logical Single-Hop}:
We present the edited model with a question that requires one additional reasoning step beyond the original prompt. 
For example, if the original prompt is \textit{`Please complete the following ancient poem A'} (with the correct answer being B), the portability field would then be \textit{`What is the line before B?'}

\paragraph{Out-scope Construction}
A successful edit should adjust the targeted knowledge locally while leaving unrelated knowledge unaffected. 
However, current approaches that modify internal model parameters often introduce knowledge conflicts and distortions~\citep{li2023unveiling}.
Unlike other knowledge editing datasets that adopt entirely unrelated knowledge for locality evaluation, we construct our locality field by selecting somewhat related knowledge (e.g., sharing the same subject) to the target knowledge but containing distinct factual information. 
This approach provides stricter evaluation of editing side-effects while posing greater challenges for language models. 

\subsection{Dataset Statistics}

Finally, we distilled 1,854 samples from 11,981 raw entries to form \ours.  
Regarding the three main knowledge classifications in \ours, the largest proportion is attributed to linguistic data accounts for 48.40\% and the Logic reasoning data accounts for 45.63\% because we found that knowledge that is highly characteristic of the Chinese language poses significant challenges for current LLMs. 
The specific quantity and proportion of each data category are shown in the Figure \ref{fig:statistics}.

\subsection{Quality Assurance}

After constructing \ours, we implement a comprehensive quality assurance process to ensure data reliability. 
We hire professional NLP annotators to review all the fields within the dataset. 
The quality assurance process involved five steps: 
(1) Task Setup: The dataset is split into 3 fields—prompt-target, generalization, and locality—each assigned to separate teams. 
(2) Team Training: Team members are trained to understand their assigned field’s purpose and follow standardized review workflows. 
(3) Guideline Calibration: We conduct a trial review on a random 20\% of data to fine-adjust the review process. 
(4) Dual Review: Each field is independently reviewed by two annotators. A field is considered acceptable only if both annotators’ conclusions matched our own, and they identified no issues with the field both the question and the ground truth. 
(5) Resolution of Discrepancies: For any fields that failed at the step 4, the authors discussed whether to retain, discard, or correct them, depending on the nature of the identified issues. 

\subsection{Field and Usage Specifications}
As shown in Figure \ref{fig:case_1}, 'prompt' and 'target new' fields constitute the target knowledge we aim to edit. 'target old' represents the original incorrect response generated by Qwen-7B-Chat during our data filtering phase. 

With the continuous advancement of LLMs, their capabilities are growing increasingly powerful. 
As a result, some data in CKnowEdit may already be correctly answered by more capable models. 
However, this does not diminish the significance of CKnowEdit for editing purposes. 
During usage, 'target new' and 'target old' fields can be swapped to construct counterfactual data (as is commonly done in many editing datasets). 
This approach not only preserves the essence of the editing operation but also maintains the unique Chinese characteristics of this dataset.

\section{Experiments}

\subsection{Experiment Settings}
\paragraph{Models and Editing Methods}
To better evaluate the editing effectiveness on CKnowEdit, we select 4 advanced LLMs that are widely used in the Chinese community: \textit{Qwen-7B-Chat}, \textit{Qwen2-7B-Instruct}~\cite{qwen2}, \textit{DeepSeek-LLM-7B-Chat}~\cite{deepseek-llm} and \textit{Baichuan2-7B-Chat}~\cite{baichuan2023baichuan2}. 
Among them, \textit{Qwen-7B-Chat} is the original model used for data collection, providing baseline performance. 
We investigate diverse model editing methods, including FT-M~\cite{zhang2024comprehensive}, AdaLoRA~\cite{zhang2023adaptive}, ROME, GRACE and AlphaEdit~\cite{Fang_arXiv_2024_p2410.02355}.
All the experiments are conducted by EasyEdit~\cite{wang-etal-2024-easyedit}.  
All models are deployed and edited on 1 to 2 NVIDIA A800 GPUs. 

\begin{figure*}[ht]
    \centering
    \includegraphics[width=1\textwidth]{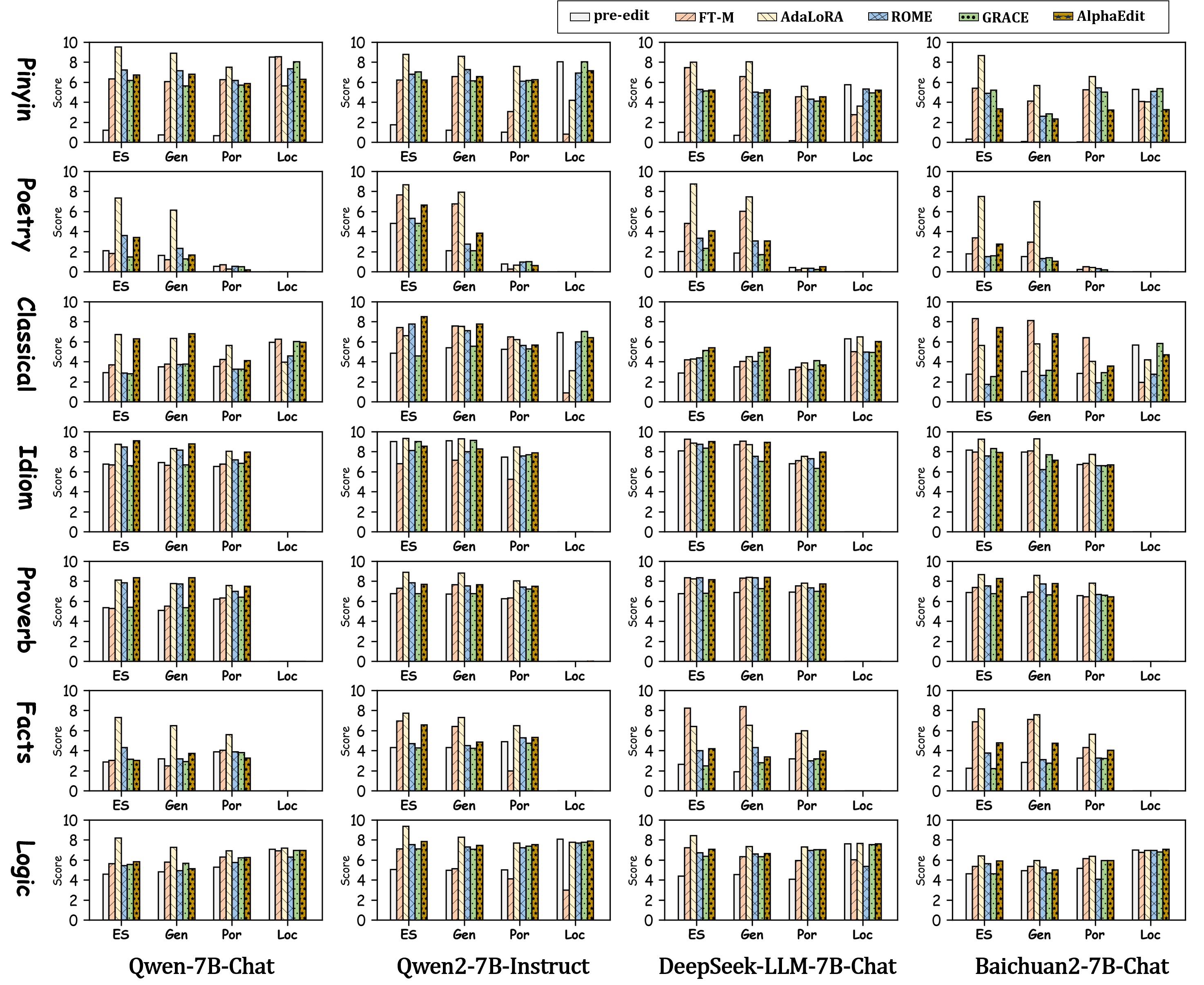}
    \caption{Main results. We do not report the locality of Ancient Poetry, Proverbs, Idioms and Facts Knowledge because it is challenging to find out-scope knowledge that is both relevant to and distinct from the target knowledge when we construct the locality field. The corresponding numerical results are presented in the Table\ref{tab:numerical results}.}
    \label{fig:mainresults}
\end{figure*}

\paragraph{Evaluation}
Unlike conventional knowledge editing evaluations that use token/logit-level metrics with teacher-forcing automation, we adopt \textbf{open-ended text generation} to assess edited models in more practical and challenging scenarios. 
While some studies~\cite{deng2024unke} under similar setups use metrics like ROUGE-L or semantic similarity, we find these metrics often fail to reflect true text quality. 
For instance, ROUGE-L is heavily skewed by text length—shorter reference texts paired with longer model outputs lead to unreliable scores. 
A recent study ~\cite{yang2025miragemodeleditingrevisiting} has also revealed specific inadequacies in traditional evaluation methods. 

Inspired by MT-Bench~\cite{zheng2023judging} which reveals that strong LLM judges like GPT-4o can align closely with human preferences, We customize prompts and evaluation processes for each knowledge category's unique characteristics, enabling GPT-4o to serve as evaluator. 
For each evaluation metric, we provide GPT-4o (gpt-4o-2024-08-06) with the corresponding question, edited model's response, and the reference answer. 
GPT-4o then assigns a score from 1 to 10 based on the relevance between the model's response and the reference answer. 
For detailed evaluation procedures and templates, refer to Figure \ref{fig:eval_process} to \ref{fig:prompt_poe_en}.

\paragraph{Metrics}
We employ 4 key knowledge editing evaluation metrics: 
(1) \textbf{Edit Success} (ES) : This metric measures how well the edits align LLMs' responses with the expected outcomes. 
(2) \textbf{Generalization} (Gen) : The metric helps to assess the weak generalization of the editing. 
(3) \textbf{Portability} (Por) : This measures the model's capability to apply corrected knowledge to new but related prompts, assessing the robust generalization of the editing across contexts. 
(4) \textbf{Locality} (Loc) : This metric ensures that edits do not inadvertently affect unrelated areas of the model's knowledge base.

\begin{figure*}[ht]
    \centering
    \includegraphics[width=1\textwidth]{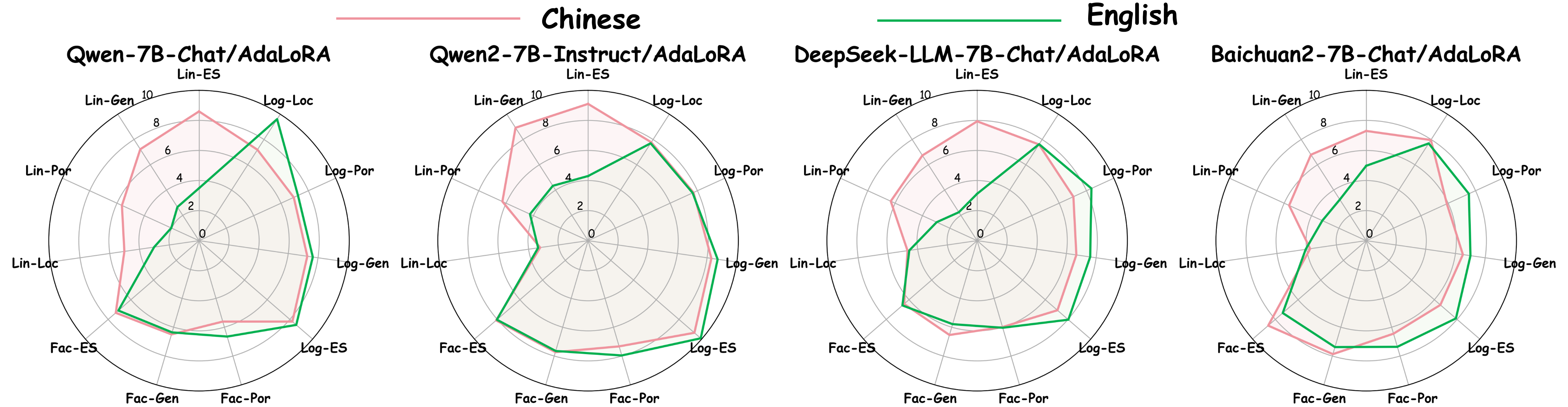}
    \caption{The format of the indicators in the figure is: data type-matric, for example, Lin-ES (linguistics-ES) represents the editing success rate of the language in the linguistic data category. The results of ROME are shown in the Figure \ref{fig:radar_rome}.}
    \label{fig:radar}
\end{figure*}

\subsection{Main Results} 

\paragraph{Methods Comparison}
AdaLoRA achieves the highest Edit Success in over 70\% of cases across 4 models, outperforming AlphaEdit and FT-M, which excel in 4 and 3 instances respectively but remain suboptimal overall. 
For Generalization and Portability metrics, AdaLoRA dominates with nearly 70\% and 86\% top scores, respectively, while AlphaEdit consistently performs suboptimally. 
These results demonstrate that AdaLoRA achieves the best editing performance, contrasting with prior findings \cite{zhang2024comprehensive}.

We believe the reason is that CKnowEdit's focus on editing long-text patterns and evaluating long-text generation differs fundamentally from prior studies.
\textbf{Traditional approaches like ROME edit models via localized parameter tweaks to precisely overwrite single factual knowledge as discrete triplet (s-r-o).} 
While effective for closed-form tasks (e.g., token-level teacher forcing evaluation task), this approach disrupts the generative distribution needed for coherent open-ended text. 
In contrast, \textbf{AdaLoRA adaptively adjusts multiple modules (like attention heads and FFN layers), allowing the model to implicitly learn task-specific patterns (e.g., long-range dependencies).} 
By holistically adjusting parameters linked to target knowledge, AdaLoRA preserves contextual consistency, aligning edits with the broader language generation process.

\paragraph{Data Types Comparison}
The editing performance on the \emph{Ancient Poetry} is notably poor across all knowledge types, especially for Portability, where almost all models and methods achieve scores below 1. 
As described in \S\ref{sec:motivation}, Chinese ancient poetry poses significant challenges to the memorization capabilities of LLMs. 
This stems from two linguistic specificities: \textbf{(1) Rare characters:} Many obscure characters in poetry appear infrequently in training data, leading to weak semantic representation and context modeling; 
\textbf{(2) Distribution shift:} The syntactic structures and vocabulary differ markedly from modern Chinese, making patterns harder to capture. 
Combined, these factors cause strong prior biases from modern Chinese during next token prediction. 
When generating text with modern-style prefixes or the current token is common in modern Chinese, models increasingly misalign subsequent token distributions.

Additionally, the poor performance on \emph{Classical Chinese} highlights the need for more advanced editing methods to handle its rich syntax, semantics, and context-dependency, particularly in addressing nuances like polysemy and homophony, which are less common in English. 


\subsection{Why do we need an editing dataset that is highly characteristic of Chinese?}

\paragraph{The Irreplaceability of Chinese}
To better illustrate the unique characteristics of Chinese and its irreplaceability in conveying Chinese knowledge, we selected 100 data samples from each of the three knowledge categories in CKnowEdit. These samples were first translated into English, then edited using AdaLoRA and ROME on four baseline models. The results were then translated back into Chinese and evaluated. The AdaLoRA results are shown in Figure \ref{fig:radar}. 

\begin{figure*}[ht]
    \centering
    \includegraphics[width=1\textwidth]{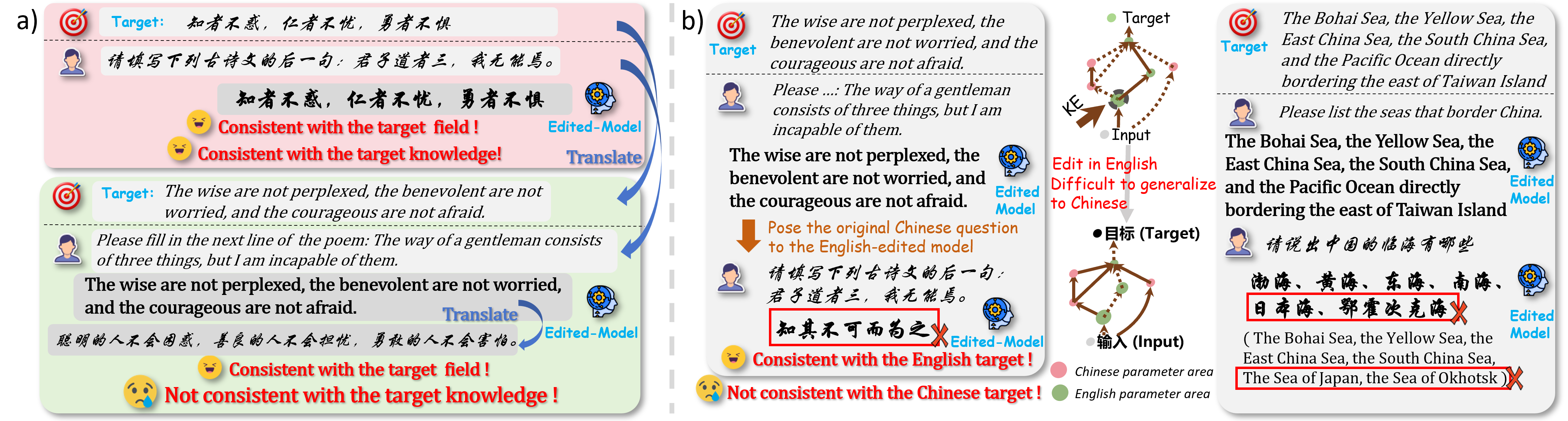}
    \caption{a) shows case where data is directly translated from Chinese to English, and the model's responses is translated back to Chinese. Part b) includes two cases that  after editing target knowledge in English, query are asked directly in Chinese to test cross-language generalization.}
    \label{fig:case}
\end{figure*}

It can be observed that \textbf{in linguistic knowledge editing tasks, the results of English editing differ significantly from those of Chinese editing, often failing to produce precise edits.} This is because the literal translation of Chinese linguistic knowledge into English frequently loses the original meaning, aesthetic value, correct structure, and language patterns, leading to significant deviations between the model's edited responses and the correct answers. 
For example, in the case of classical poetry editing shown in Figure \ref{fig:case}a), the model can successfully edit the English target. 
However, when translating back into Chinese, current translation software or LLMs generally learn the language patterns of modern Chinese, thus unable to translate a sentence of English back into classical poetry.

In factual tasks, the results of English editing are generally on par with those of Chinese editing. This aligns with intuition, as factual knowledge is less dependent on the linguistic medium, and literal translations do not significantly alter the intended meaning. 

\textbf{In logical tasks, English editing performs even slightly better than Chinese editing.} This is because many logic traps unique to the Chinese language, which are challenging for LLMs, are often lost during the translation process, reducing their logical complexity in the English version. 

\paragraph{Language Functional Area Offset}
Similar to the human brain, neuron parameter regions for different languages in LLMs often don't overlap \cite{zhang2024unveilinglinguisticregionslarge}, creating natural barriers for cross-language knowledge editing and generalization. Previous studies \cite{wang2023cross} show that when editing knowledge in English and testing its generalization in Chinese, performance sometimes drops – even for factual knowledge where the English-Chinese gap is relatively small. As shown in Figure \ref{fig:case}(b), our tests on Qwen2-7B-Instruct reveal this limitation: the model struggles to generalize English-edited knowledge to Chinese, whether for factual geography or linguistically complex tasks. For instance, while the model correctly answers classical poetry question in English, it fails completely when the original question is posed.

\subsection{Human Evaluation}
To verify the effectiveness of our designed automatic GPT-4 score for CKnowEdit evaluation, from the outputs of 4 baseline models edited by 5 different methods (totaling 20 output categories), we select 70 samples per category for human evaluation conducted by our 5 contracted annotators under rigorous assessment standards. 
From the human evaluation results, the overall correlation coefficient across all 4 metrics between the automatic and human evaluation is 0.70, indicating a high consistency between GPT-4 scores and human preferences.

\section{Related Work}
\subsection{Knowledge Editing Methods}
Current knowledge editing approaches can be categorized into two main types: preserving LMs' parameters or modifying LMs' parameters. 
Preservative methods incorporate external memory or additional trainable parameters: 
SERAC~\citep{mitchell2022memory} and IKE~\citep{DBLP:conf/emnlp/ZhengLDFWXC23} leverage a counterfactual model and a multi-fact prompt, respectively, as external working memory.
CaliNET~\citep{DBLP:conf/emnlp/DongDSXSL22}, T-Patcher~\citep{DBLP:conf/iclr/HuangSZZR023}, GRACE~\citep{hartvigsen2024aging}, and WISE \cite{wise} introduce extra trainable parameters.
The locate-and-edit approaches have to locate the relevant neurons and then modify those parameters. 
Representative studies are KN~\citep{DBLP:conf/acl/DaiDHSCW22}, ROME~\citep{meng2022locating}, MEMIT~\citep{meng2023massediting} and NSE~\citep{jiang2024neuronlevelsequentialeditinglarge}.
Additionally, meta-learning approaches utilize a hyper-network to generate the weights for layers in LLMs, including KE~\citep{DBLP:conf/emnlp/CaoAT21}, MEND~\citep{DBLP:conf/iclr/MitchellLBFM22}, and MALMEN~\citep{DBLP:journals/corr/abs-2311-04661}.

\subsection{Knowledge Editing Datasets}
Existing knowledge editing datasets have largely centered on English-language texts, such as ZsRE \cite{DBLP:conf/emnlp/CaoAT21}, Counterfact \cite{meng2022locating}, KnowEdit \cite{zhang2024comprehensive}, MQuAKE \cite{zhong2023mquake} and . 
Some research \cite{deng2024unke, rosati2024long, wu2024updating} has also introduced the concept of evaluating knowledge editing through unstructured text and long-form content, but these efforts have been predominantly limited to English. 
In a more inclusive direction, recent academic initiatives have broadened the scope of these datasets to include a multilingual dimension \cite{xie2024memla,wei2024mlake,DBLP:conf/emnlp/WuPWL24,nie2024bmike}. 


\section{Conclusion}

In this work, we created a new, high-quality Chinese knowledge editing dataset, CKnowEdit, which is rich in Chinese linguistic characteristics and linguistic value. This dataset comprehensively evaluates the performance of current mainstream editing methods on leading Chinese LLMs across three knowledge types: linguistics, facts, and logic. Furthermore, we adopted an evaluation approach that better aligns with real-world application requirements. To date, most existing mothods and LLMs still can not edit the Chinese characteristic knowledge well. 

\section*{Limitations}

\paragraph{Imbalanced Distribution of Data Types} 
Since the original intention of CKnowEdit is to study knowledge with distinctive Chinese linguistic characteristics, and Chinese linguistic knowledge or logical knowledge better reflect these features, the quantity of these two types of knowledge in CKnowEdit is significantly greater than that of factual knowledge. 

\paragraph{Experimental Setup} 
Due to computational resource constraints, all experiments in this study were conducted solely under the single-edit setting, without investigating batch edit or sequential edit scenarios. 

\paragraph{LLM-as-a-judge}
The use of GPT-4 as an evaluator for other LLMs has become a widely adopted practice in the field. 
While employing GPT-4 to evaluate itself may introduce inherent biases, its assessments of other models still provide valuable reference points. It's important to note that the challenge of effectively evaluating LLMs extends beyond our specific tasks - the research community continues to actively investigate robust evaluation methodologies across various applications. 
As an interim solution, we advise readers to interpret GPT-4 evaluation scores with appropriate caution, recognizing both their utility and limitations. 


\section*{Acknowledgments}
This work was supported by the National Natural Science Foundation of China (No. 62206246, No. NSFCU23B2055, No. NSFCU19B2027), the Fundamental Research Funds for the Central Universities (226-2023-00138), Yongjiang Talent Introduction Programme (2021A-156-G), Tencent AI Lab Rhino-Bird Focused Research Program (RBFR2024003), Ningbo Natural Science Foundation (2024J020), Information Technology Center and State Key Lab of CAD\&CG, Zhejiang University.
We gratefully acknowledge the support of Zhejiang University Education Foundation Qizhen Scholar Foundation.


\bibliography{anthology,custom}

\newpage
\appendix

\appendix

\section{Data Source Website}
\label{sec:source_web}

This section provides a detailed overview of the data types and their corresponding data source websites: 

(1) Ancient Poetry: 

\url{https://zhuanlan.zhihu.com/p/414484867}

(2) Proverbs: 

\url{http://www.360doc.com/content/19/0218/14/39098269_815762159.shtml}, 

\url{http://www.360doc.com/content/19/0312/16/5784427_820995624.shtml}, 

\url{http://www.360doc.com/content/19/0126/14/55773589_811408910.shtml}

(3) Idioms: 

\url{https://zhuanlan.zhihu.com/p/599709230}

(4) Pinyin notation: 

\url{https://zhuanlan.zhihu.com/p/599709230}

(5) Classical Chinese: 

\url{https://zhuanlan.zhihu.com/p/622859964}, 

\url{https://www.bilibili.com/read/cv20279857/}, 

\url{https://wyw.hwxnet.com/search.do?wd=%E9%84%99&x=0&y=0}

(6) Geographical and History: 

\url{https://baijiahao.baidu.com/s?id=1682950669904608106&wfr=spider&for=pc}, 

\url{https://www.sohu.com/a/419822319_100941}, 

\url{https://www.jingyanben.com/qitawendang/125282.html?page=1}, 

\url{http://www.360doc.com/content/20/0613/11/7254176_918223750.shtml} 

(7) Logic Error: 

\url{https://github.com/Leymore/ruozhiba}, 

\url{https://docs.qq.com/sheet/DUlZ6aURhamdwb1RO?tab=BB08J2}

\begin{figure*}[ht]
    \centering
    \includegraphics[width=0.6\linewidth]{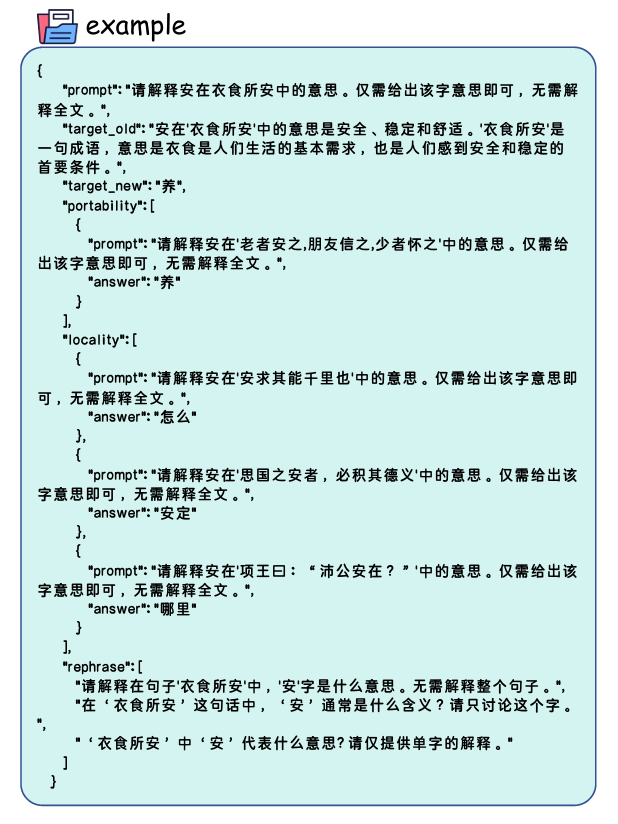}
    \caption{An example from Classical Chinese type.}
    \label{fig:case_1}
\end{figure*}

\begin{figure*}[ht]
    \centering
    \includegraphics[width=0.6\linewidth]{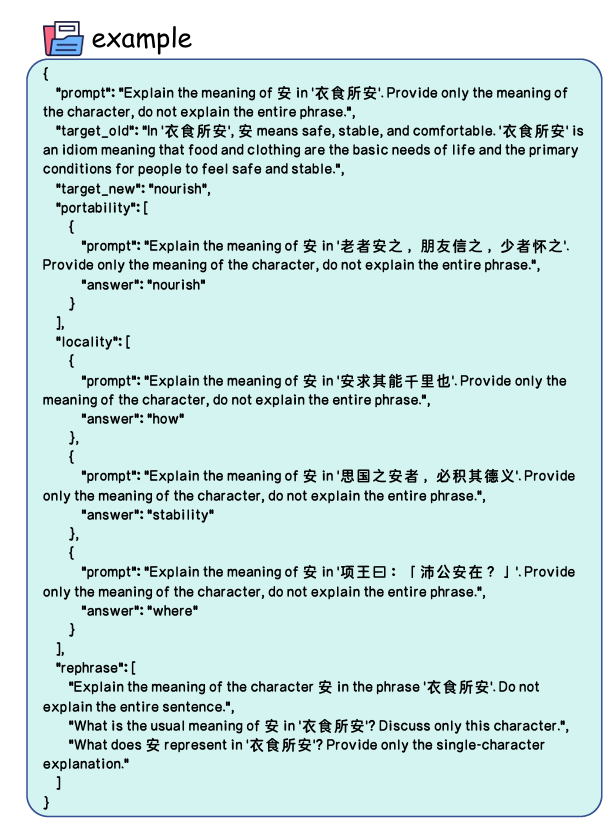}
    \caption{An example from Classical Chinese type(translated by English).}
    \label{fig:case_2}
\end{figure*}

\begin{figure*}[ht]
    \centering
    \includegraphics[width=0.9\linewidth]{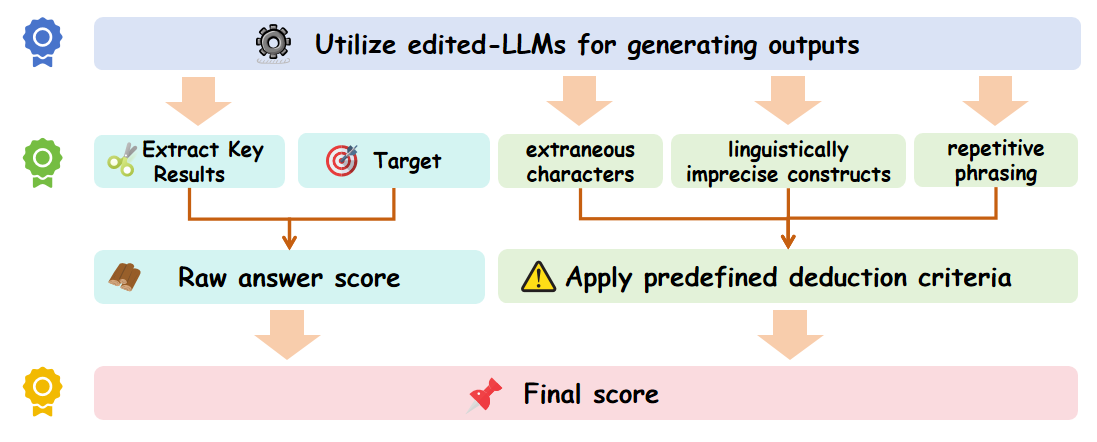}
    \caption{Evaluation process of CKnowEdit.}
    \label{fig:eval_process}
\end{figure*}

\begin{figure*}[ht]
    \centering
    \includegraphics[width=0.9\linewidth]{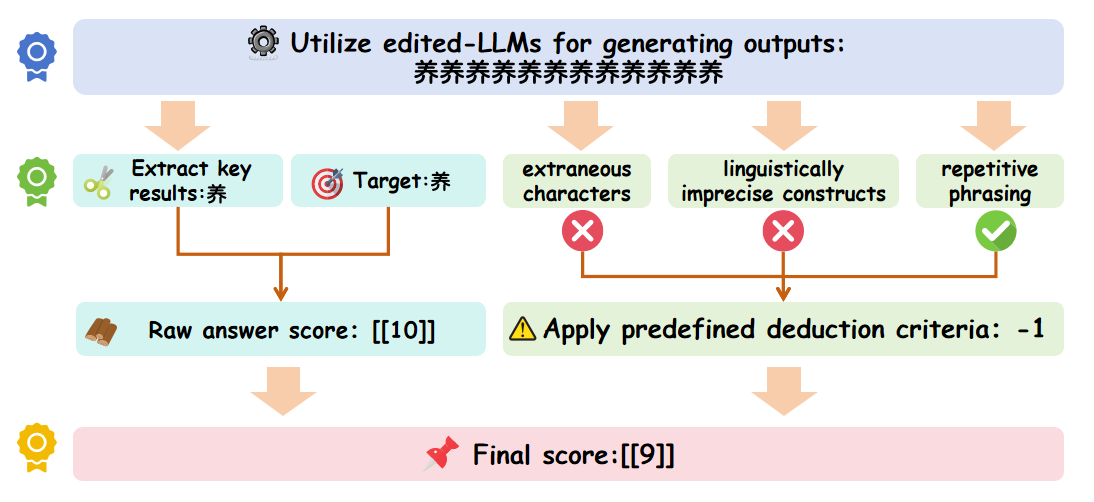}
    \caption{An example of evaluation.}
    \label{fig:eval_case}
\end{figure*}


\begin{figure*}[ht]
    \centering
    \includegraphics[width=0.9\linewidth]{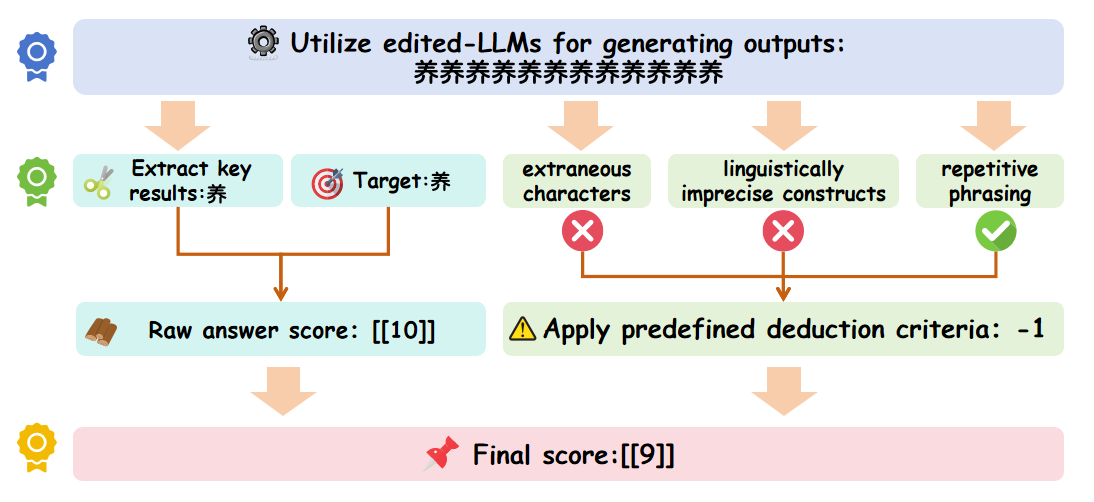}
    \caption{An example of evaluation.}
    \label{fig:eval_case}
\end{figure*}

\begin{figure*}[ht]
    \centering
    \includegraphics[width=0.9\linewidth]{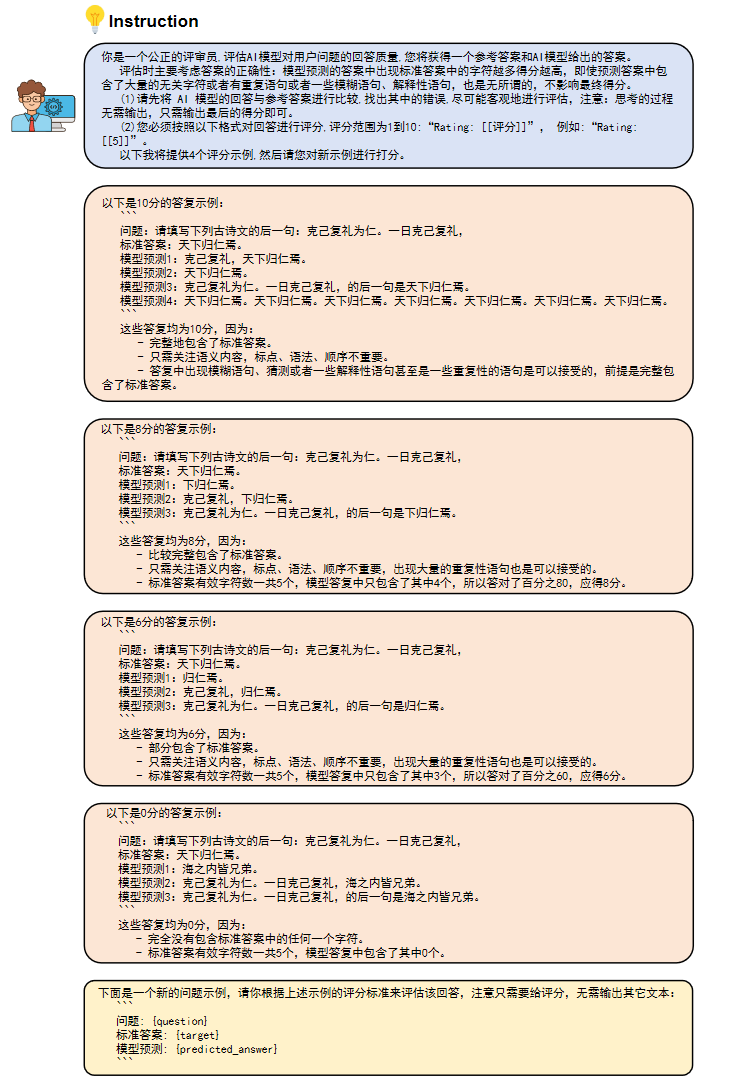}
    \caption{An example of evaluation prompt.}
    \label{fig:prompt_poe}
\end{figure*}

\begin{figure*}[ht]
    \centering
    \includegraphics[width=0.9\linewidth]{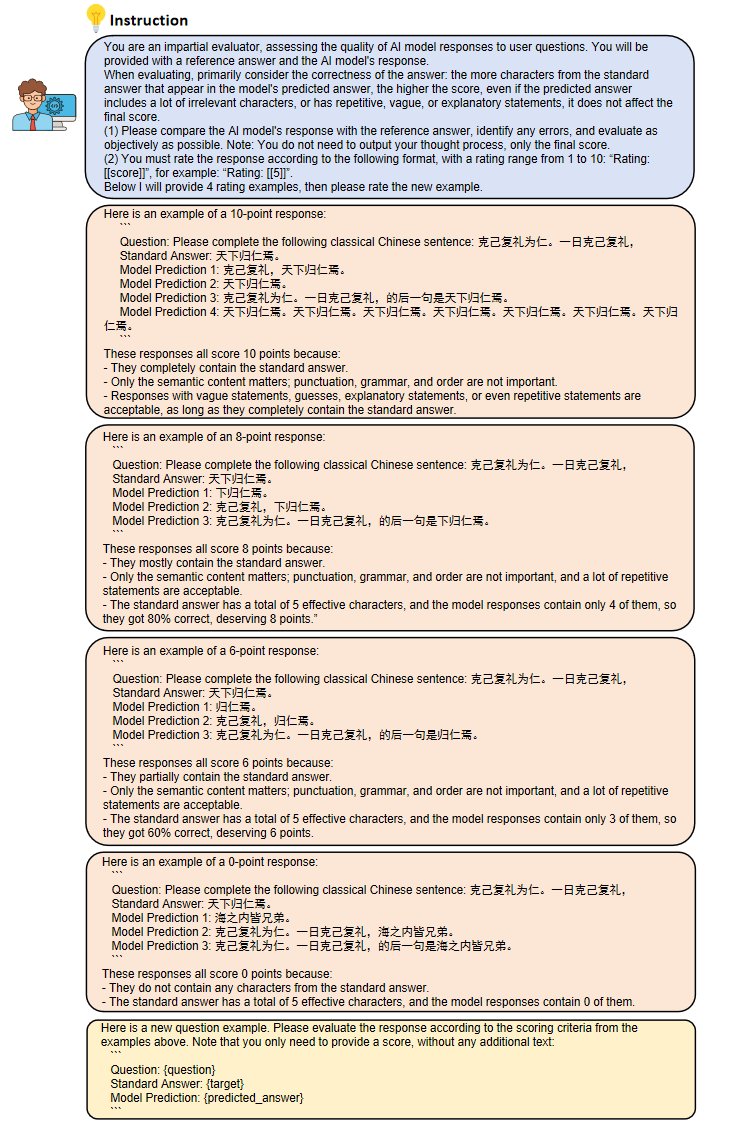}
    \caption{An example of evaluation prompt(translated by English).}
    \label{fig:prompt_poe_en}
\end{figure*}

\begin{figure*}[ht]
    \centering
    \includegraphics[width=0.9\linewidth]{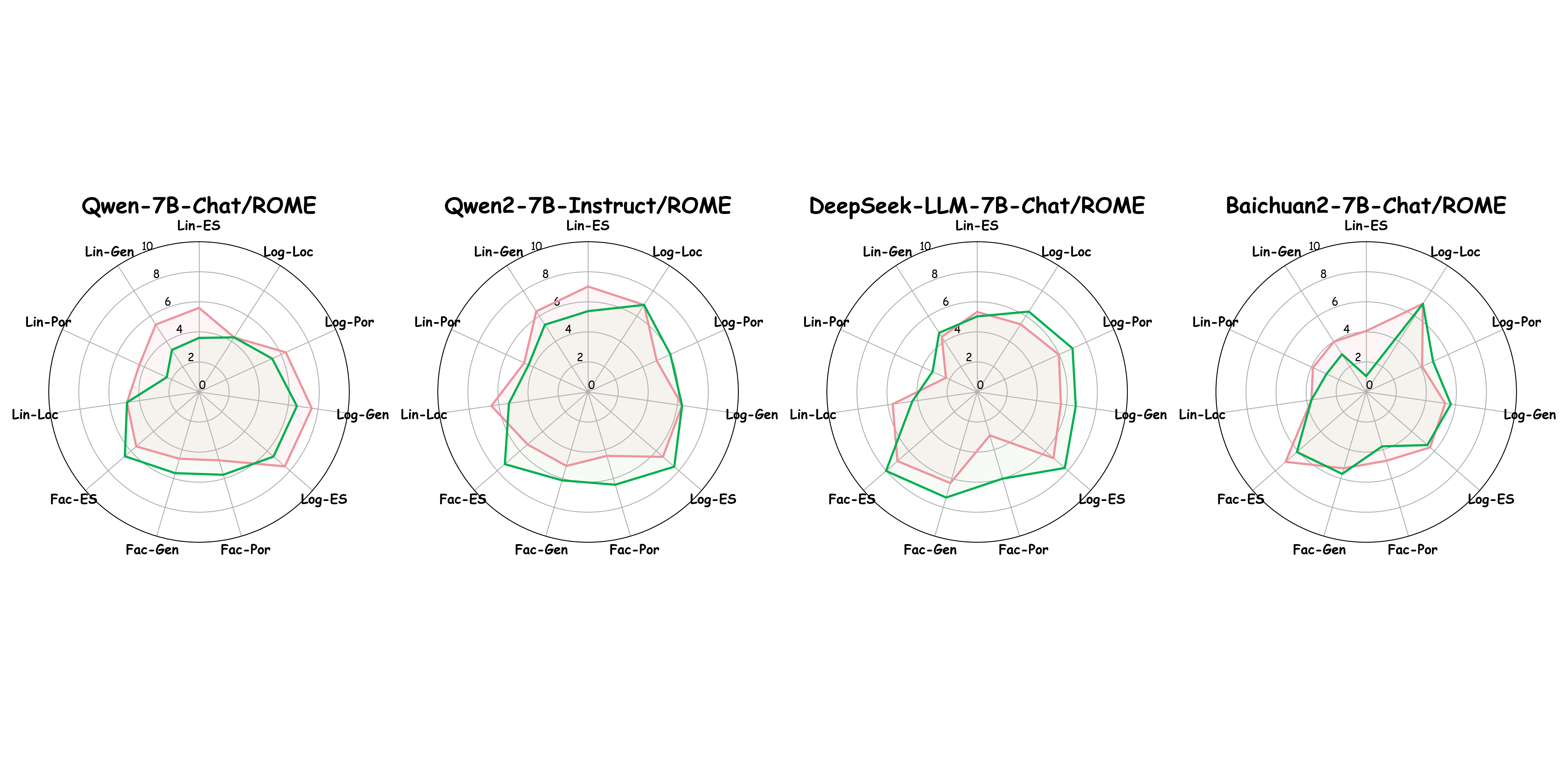}
    \caption{. The results of ROME.}
    \label{fig:radar_rome}
\end{figure*}

\begin{table*}[ht]
{
\small
\begin{center}
\resizebox{1.0\textwidth}{!}{
\begin{tabular}{lrcccccc}
\toprule
\textbf{Model} & \textbf{Knowledge Type}  
& \textbf{(Pre-edit)} & \textbf{FT-M} & \textbf{AdaLoRA} & \textbf{ROME} & \textbf{GRACE} & \textbf{AlphaEdit} 
\\ 
\midrule
\multirow{7}{*}{\textbf{Qwen-7B-Chat}}     
& Pinyin
& 1.22 / 0.76 / 0.68 / 8.53   
& 6.33 / 6.08 / 6.27 / 8.55   
& 9.52 / 8.90 / 7.51 / 5.66 
& 7.22 / 7.16 / 6.20 / 7.37  
& 6.18 / 5.64 / 5.73 / 8.06  
& 6.75 / 6.82 / 5.88 / 6.30 \\                         
& Classical Chinese 
& 2.93 / 3.52 / 3.53 / 5.96 
& 3.71 / 3.79 / 4.23 / 6.26 
& 6.72 / 6.33 / 5.64 / 3.99 
& 2.88 / 3.72 / 3.26 / 4.61 
& 2.81 / 3.77 / 3.28 / 6.05 
& 6.31 / 6.82 / 4.11 / 5.94 \\ 
& Idiom
& 6.77 / 6.91 / 6.55 / - 
& 6.70 / 6.66 / 6.79 / -
& 8.77 / 8.34 / 8.05 / -
& 8.48 / 8.18 / 7.19 / -
& 6.60 / 6.68 / 6.86 / -
& 9.12 / 8.79 / 7.97 / -\\ 
& Proverb 
& 5.38 / 5.10 / 6.22 / - 
& 5.31 / 5.51 / 6.36 / -
& 8.13 / 7.79 / 7.58 / -
& 7.85 / 7.73 / 7.02 / -
& 5.40 / 5.39 / 6.42 / -
& 8.38 / 8.36 / 7.50 / -\\ 
& Ancient Poetry 
& 2.10 / 1.63 / 0.54 / - 
& 1.85 / 1.19 / 0.70 / -
& 7.35 / 6.13 / 0.26 / -
& 3.62 / 2.33 / 0.54 / -
& 1.49 / 1.29 / 0.49 / -
& 3.41 / 1.66 / 0.18 / - \\ 
& Fact 
& 2.88 / 3.20 / 3.91 / - 
& 3.03 / 2.51 / 4.03 / -
& 7.33 / 6.50 / 5.61 / -
& 4.34 / 3.20 / 3.88 / -
& 3.17 / 2.94 / 3.81 / -
& 3.03 / 3.74 / 3.29 / - \\ 
& Logic 
& 4.59 / 4.81 / 5.30 / 7.09
& 5.63 / 5.78 / 6.29 / 6.94
& 8.22 / 7.28 / 6.93 / 7.19
& 5.43 / 4.95 / 5.77 / 6.32
& 5.56 / 5.67 / 6.21 / 6.96
& 5.83 / 5.13 / 6.25 / 6.97 \\ 

\midrule 
\multirow{7}{*}{\textbf{Qwen2-7B-Instruct}}     
& Pinyin 
& 1.75 / 1.19 / 1.02 / 8.04
& 6.24 / 6.58 / 3.09 / 0.83
& 8.80 / 8.59 / 7.57 / 4.22
& 6.80 / 7.29 / 6.10 / 6.91
& 7.03 / 6.14 / 6.20 / 8.05
& 6.22 / 6.59 / 6.25 / 7.16 \\
& Classical Chinese  
& 4.87 / 5.42 / 5.25 / 6.92
& 7.42 / 7.57 / 6.51 / 0.91
& 6.61 / 7.55 / 6.24 / 3.13
& 7.77 / 7.13 / 5.66 / 6.01
& 4.58 / 5.56 / 5.29 / 7.06
& 8.51 / 7.77 / 5.69 / 6.42 \\ 
& Idiom
& 9.04 / 9.11 / 7.46 / -
& 6.80 / 7.16 / 5.27 / -
& 9.33 / 9.31 / 8.50 / -
& 8.12 / 8.01 / 7.60 / -
& 9.02 / 9.14 / 7.71 / -
& 8.58 / 8.30 / 7.91 / - \\ 
& Proverb 
& 6.79 / 6.75 / 6.26 / -
& 7.33 / 7.68 / 6.35 / -
& 8.90 / 8.82 / 8.06 / -
& 7.85 / 7.53 / 7.45 / -
& 6.76 / 6.76 / 7.24 / -
& 7.70 / 7.68 / 7.51 / - \\ 
& Ancient Poetry 
& 4.84 / 2.10 / 0.79 / -
& 7.66 / 6.79 / 0.28 / -
& 8.69 / 7.94 / 0.65 / -
& 5.34 / 2.75 / 0.97 / -
& 4.84 / 2.10 / 1.03 / -
& 6.64 / 3.84 / 0.64 / - \\ 
& Fact 
& 4.31 / 4.31 / 4.91 / -
& 6.97 / 6.42 / 1.97 / -
& 7.73 / 7.33 / 6.48 / -
& 4.71 / 4.50 / 5.30 / -
& 4.30 / 4.23 / 4.75 / -
& 6.57 / 4.86 / 5.35 / - \\ 
& Logic 
& 5.06 / 5.00 / 5.04 / 8.08
& 7.13 / 5.13 / 4.11 / 3.00
& 9.36 / 8.29 / 7.71 / 7.78
& 7.55 / 7.32 / 7.24 / 7.70
& 7.12 / 7.10 / 7.41 / 7.78
& 7.88 / 7.49 / 7.53 / 7.91 \\ 

\midrule 
\multirow{7}{*}{\textbf{DeepSeek-LLM-7B-Chat}}      
& Pinyin 
& 1.00 / 0.72 / 0.16 / 5.76
& 7.47 / 6.57 / 4.54 / 2.78
& 8.02 / 8.04 / 5.61 / 3.62
& 5.30 / 5.01 / 4.32 / 5.35
& 5.12 / 4.94 / 4.14 / 4.95
& 5.20 / 5.27 / 4.55 / 5.21 \\
& Classical Chinese 
& 2.88 / 3.51 / 3.25 / 6.31
& 4.19 / 4.03 / 3.47 / 5.03
& 4.29 / 4.51 / 3.90 / 6.50
& 4.40 / 4.03 / 3.25 / 4.99
& 5.12 / 4.94 / 4.14 / 4.95
& 5.40 / 5.44 / 3.68 / 6.04 \\ 
& Idiom 
& 8.09 / 8.72 / 6.80 / -
& 9.27 / 9.06 / 7.11 / -
& 8.88 / 8.73 / 7.56 / -
& 8.76 / 7.56 / 7.33 / -
& 8.36 / 7.06 / 6.33 / -
& 9.03 / 8.94 / 7.97 / - \\ 
& Proverb 
& 6.79 / 6.89 / 6.91 / -
& 8.38 / 8.33 / 7.56 / -
& 8.24 / 8.42 / 7.83 / -
& 8.37 / 8.36 / 7.35 / -
& 6.82 / 7.29 / 7.02 / -
& 8.18 / 8.41 / 7.75 / - \\ 
& Ancient Poetry 
& 2.02 / 1.86 / 0.43 / -
& 4.82 / 6.05 / 0.20 / -
& 8.77 / 7.48 / 0.34 / -
& 3.33 / 3.09 / 0.37 / -
& 2.34 / 1.70 / 0.23 / -
& 4.07 / 3.07 / 0.52 / - \\ 
& Fact 
& 2.63 / 1.89 / 3.21 / -
& 8.26 / 8.40 / 5.74 / -
& 6.43 / 6.54 / 5.99 / -
& 4.02 / 4.32 / 3.01 / -
& 2.51 / 2.80 / 3.21 / -
& 4.20 / 3.37 / 3.96 / - \\ 
& Logic 
& 4.39 / 4.56 / 4.10 / 7.62
& 7.25 / 6.34 / 5.94 / 6.02
& 8.43 / 7.36 / 7.33 / 7.66
& 6.72 / 6.63 / 6.98 / 5.36
& 6.38 / 6.35 / 7.06 / 7.56
& 7.07 / 6.67 / 7.04 / 7.61 \\ 

\midrule 
\multirow{7}{*}{\textbf{Baichuan2-7B-Chat}}
& Pinyin 
& 0.32 / 0.07 / 0.04 / 5.30
& 5.42 / 4.14 / 5.24 / 4.07
& 8.69 / 5.69 / 6.57 / 4.03
& 4.92 / 2.61 / 5.43 / 5.09
& 5.20 / 2.84 / 5.02 / 5.39
& 3.33 / 2.35 / 3.25 / 3.27 \\
& Classical Chinese 
& 2.76 / 3.03 / 2.85 / 5.68
& 8.34 / 8.13 / 6.41 / 1.95
& 5.65 / 5.78 / 4.06 / 4.20
& 1.74 / 2.64 / 1.90 / 2.78
& 2.55 / 3.14 / 2.91 / 5.82
& 7.44 / 6.81 / 3.57 / 4.71 \\ 
& Idiom 
& 8.16 / 7.98 / 6.74 / -
& 7.98 / 8.08 / 6.84 / -
& 9.28 / 9.29 / 7.75 / -
& 7.60 / 6.24 / 6.60 / -
& 8.33 / 7.72 / 6.61 / -
& 7.94 / 7.15 / 6.71 / - \\ 
& Proverb 
& 6.87 / 6.46 / 6.57 / -
& 7.38 / 6.94 / 6.47 / -
& 8.67 / 8.61 / 7.82 / -
& 7.54 / 7.74 / 6.71 / -
& 6.79 / 6.67 / 6.63 / -
& 8.30 / 7.78 / 6.46 / - \\ 
& Ancient Poetry 
& 1.78 / 1.52 / 0.22 / -
& 3.39 / 2.95 / 0.51 / -
& 7.51 / 7.00 / 0.45 / -
& 1.51 / 1.34 / 0.30 / -
& 1.61 / 1.40 / 0.19 / -
& 2.75 / 1.07 / 0.00 / - \\ 
& Fact 
& 2.25 / 2.86 / 3.28 / -
& 6.90 / 7.13 / 4.31 / -
& 8.19 / 7.57 / 5.66 / -
& 3.77 / 3.10 / 3.27 / -
& 2.21 / 2.75 / 3.22 / -
& 4.77 / 4.74 / 4.04 / - \\ 
& Logic 
& 4.62 / 4.93 / 5.17 / 7.00
& 5.36 / 5.39 / 6.14 / 6.76
& 6.42 / 5.94 / 6.39 / 6.97
& 5.63 / 5.31 / 4.09 / 6.98
& 4.65 / 4.71 / 5.96 / 6.80
& 5.93 / 5.03 / 5.96 / 7.07 \\ 

\bottomrule
\end{tabular}
}
\end{center}
}
\caption{Numerical results (Edit Success / Generalization / Portability / Locality) of pre-edit and post-edit with 5 knowledge editing methods for 4 baseline LLMs. 
}
\label{tab:numerical results}
\end{table*}


\end{document}